\documentclass[conference]{IEEEtran}
\usepackage{times}

\usepackage[numbers]{natbib}
\usepackage{multicol}
\usepackage[bookmarks=true]{hyperref}

\usepackage{color}
\usepackage{comment}
\usepackage{graphicx}
\usepackage{verbatim} 
\usepackage{subcaption} 
\usepackage{float} 
\usepackage{url}
\usepackage{listings}
\usepackage{balance}

\begin{document}

\title{Towards Probabilistic Planning of Explanations for Robot Navigation}

\author{Author Names Omitted for Anonymous Review. Paper-ID [add your ID here]}

\author{
\authorblockN{Amar Halilovic}
\authorblockA{Institute of Artificial Intelligence\\
Ulm University\\
Ulm, Germany\\
Email: amar.halilovic@uni-ulm.de}
\and
\authorblockN{Senka Krivic}
\authorblockA{Faculty of Electrical Engineering\\
University of Sarajevo\\
Sarajevo, BiH\\
Email: senka.krivic@etf.unsa.ba}
\and
\authorblockN{Gerard Canal}
\authorblockA{Department of Informatics\\ 
King’s College London\\
London, UK\\
Email: gerard.canal@kcl.ac.uk}}


%

\maketitle

\begin{abstract}
In robotics, ensuring that autonomous systems are comprehensible and accountable to users is essential for effective human-robot interaction. This paper introduces a novel approach that integrates user-centered design principles directly into the core of robot path planning processes. We propose a probabilistic framework for automated planning of explanations for robot navigation, where the preferences of different users regarding explanations are probabilistically modeled to tailor the stochasticity of the real-world human-robot interaction and the communication of a robot’s decisions and actions towards humans. This approach aims to enhance the transparency of robot path planning and adapt to diverse user explanation needs by anticipating the types of explanations that will satisfy individual users. 
\end{abstract}

\IEEEpeerreviewmaketitle

\section{Introduction}
The lack of transparency in robot decision-making~\cite{edmonds2019tale,wachter2017transparent} impedes the faster integration of robots into human society. 
To facilitate this integration, robots must exhibit characteristics reminiscent of social robots~\cite{breazeal2016social}.
Social robots generally fall into the categories of service (utilitarian) or assistive (affective)~\cite{baraka2020extended} robots.
Thus, they should respect social norms of human interaction and behavior.
Despite the challenges in acquiring such social norms, the presence of robots has increased rapidly, thus escalating the demand for explicability in robot actions~\cite{lim2021social}.

Offering explanations for robot actions has been shown to yield favorable effects on human trust~\cite{leichtmann2023effects} and comprehension~\cite{van2014explaining}. 
Moreover, it is pivotal in nurturing effective human-robot interaction (HRI)~\cite{setchi2020explainable}. 
An explainable robot is also perceived as more socially adept~\cite{ambsdorf2022explain}.
Despite the manifold advantages that robots bring, a deficiency in transparency and accountability persists concerning their decision-making processes~\cite{felzmann2019robots}. 
The complexity of most robot behaviors and human-robot social interactions augments this challenge.

We constitute a problem of explainability in robotics in the context of robot path planning in social settings, i.e., social navigation. Imagine a navigating robot experiencing a planner failure without successful recovery behavior. Such a robot could cause harm to people around it, depending on the environment and navigation context. To justify its behavior and mitigate negative consequences, the robot should be able to explain the underlying reasons for its behavior.
However, different people, i.e., explanation recipients (explainees), may have different preferences regarding explanations. The robot must be able to judge how to characterize its explanations based on the preferences of the explainees. To give robots this ability, we conceptualize generating explanations as a probabilistic planning problem, where explanations, conditioned on user explanation preferences, can be planned as other robot actions, e.g., joint motion. 

\section{AI Planning}
Automated Planning and Scheduling, commonly known as AI Planning, is a subfield of Artificial Intelligence (AI) dedicated to devising strategies for achieving specified objectives through a sequence of actions. It involves formalizing planning problems, where an AI system must decide on a set of actions to transition from an initial state to a goal state, considering constraints and available resources. A cornerstone in AI Planning is the Planning Domain Definition Language (PDDL), introduced in 1998~\cite{aeronautiques1998pddl}, which provides a standardized way to represent planning problems. PDDL has been widely adopted in various domains, such as robotics, logistics, and space exploration. For example, NASA's Mars rovers have used PDDL to plan and execute daily activities autonomously, demonstrating the practicality and robustness of discrete planning in managing complex and dynamic tasks in unknown environments.

Probabilistic planning extends the traditional discrete planning framework by incorporating uncertainty and stochastic elements into decision-making. This is effectively modeled using the Relational Dynamic Influence Diagram Language (RDDL), developed by Sanner~\cite{sanner2010relational} in 2010. RDDL allows for the representation of probabilistic effects and relational structures, making it well-suited for environments where actions have uncertain outcomes. Probabilistic planning is particularly valuable in medical treatment planning, where patient responses to treatments are inherently uncertain, or in autonomous driving, where environmental conditions and other drivers' behaviors can be unpredictable. However, while probabilistic planning offers a more nuanced and realistic approach, it has significant computational challenges. The need to account for multiple possible outcomes increases the complexity of planning algorithms, often requiring more computational resources and sophisticated techniques to find optimal solutions. Despite these challenges, the ability to reason under uncertainty makes probabilistic planning indispensable for creating adaptive and resilient AI systems capable of operating effectively in the real world.


\section{Related Work}
Explainable AI Planning (XAIP) in robotics has gained attention recently.
Cashmore et al.~\cite{cashmore2019towards} have laid the groundwork by integrating classical planning techniques with explainability frameworks to make robotic actions more interpretable to human users.
A few studies have focused on interactive systems where robots can justify their actions and decision-making processes in real time---a crucial advancement for collaborative human-robot environments. For instance, research by Sreedharan et al.~\cite{sreedharan2020bridging} has introduced approaches for producing contrastive explanations that help users understand why a robot chose one plan over another. Additionally, incorporating user feedback to refine and improve the explainability of robotic actions~\cite{chakraborti2019plan} represents a significant step towards creating more intuitive and user-friendly robotic systems.

In motion and path planning literature, explanations encompass failure instances and contrastive scenarios. Failure explanations explain the cause of failure, while contrastive explanations explain why a planner selects trajectory A over an expected trajectory B.
User explanation queries regarding plans typically take a contrasting form, such as ``Why A over B?"~\cite{krarup2021contrastive}. 
Brandao et al.~\cite{brandao2021experts,brandao2021towards} present a preliminary taxonomy of explainable motion planning techniques.
Furthermore, they introduce two explainable motion planning strategies---optimization-based and sampling-based---capable of addressing both failure and contrastive questions. Their strategies imply using inherently explainable motion planners or making existing planners inherently explainable.
Rosenthal et al.~\cite{rosenthal2016verbalization} introduced a verbalization strategy for robots to provide verbal explanations. 

Although there is some work on explainable automated planning, there is only one work on automated explanation planning to our knowledge.
Halilovic and Krivic~\cite{halilovic2024planning} introduce deterministic planning of explanations using Planning Domain Definition Language (PDDL) 2.1~\cite{fox2003pddl2}. Using the deterministic nature of PDDL, they showed that the planning of explanation is possible, along with other robot actions.  However, they disregard the stochastic nature of social interaction and the uncertainty in robot reasoning of human explanation preferences.
We model the robot's reasoning about human explanation preferences as a probabilistic automated planning problem.
The planning aspect of our explanations involves the content, timing, and modality aspects, ensuring that explanations are provided when most beneficial to the user.

\section{Probabilistic Planning of Explanations}
RDDL is a powerful formalism for modeling decision-theoretic planning problems involving stochastic dynamics and complex relational structures. When modeling human explanation preferences in RDDL, we can treat these preferences as probabilistic variables that influence the robot's decisions regarding explanation representation, detail level, duration, and scope.
To model human explanation preferences as probabilistic variables in RDDL, we define several key components:

\subsection*{State Variables}
State variables represent the current state of the robot and the human user's preferences. Let \textit{S} be the set of state variables, including the robot's state $S_r$ =  \{$s_{r1},s_{r2},...,s_{rn}$\} and human preferences $S_h$ =  \{$s_{h1},s_{h2},...,s_{hm}$\}.
Human preferences can be represented as probabilistic variables $P_h$, which include preferences for explanation representation $P_r$, detail level $P_{dl}$, duration $P_d$, and scope $P_s$: $P_h$ =  \{$P_r,P_{dl},P_d,P_s$\}.

\subsection*{Action Variables}
Action variables represent the possible actions the robot can take, including providing explanations. Let \textit{A} be the set of action variables, including actions related to explanations $A$ =  \{$a_{explain},a_{move},a_{task}$\}.

\subsection*{Reward Functions}
Reward functions capture the utility of different outcomes based on how well the robot's explanations meet the human's preferences. The reward function \textit{R} reflects the satisfaction of human preferences and the successful completion of tasks: $R(s,a)$ is a function of the state $s \in S$ and action $a \in A$, incorporating human preference variables $P_h$.   

\subsection*{Transition Functions}
Transition functions define how state variables evolve based on actions and probabilistic effects. The transition function \textit{T} defines the probability of moving from one state to another, considering the probabilistic nature of human preferences: $T(s'|s,a)$ where \textit{s'} is the next state, \textit{s} is the current state, and \textit{a} is the action taken.

\subsection*{Explanation Representation}
Let $P_s$ be a Bernoulli random variable representing the probability that the user prefers explanations of a particular representation (textual, visual): $P_r$ = $P(E_r = textual)$, where $E_r$ is an indicator variable (1 if the user prefers a textual explanation, 0 otherwise).

\subsection*{Explanation Detail Level}
Let $P_{dl}$ be a Bernoulli random variable representing the probability that the user prefers explanations of a certain detail level \textit{t}: $P_{dl}$ = $P(E_{dl} = rich)$, where $E_{dl}$ is an indicator variable (1 if the user prefers an explanation with a rich level of detail, 0 otherwise).

\subsection*{Explanation Duration}
Let $P_d$ be a Bernoulli random variable representing the probability that the user prefers explanations of a particular duration (short, long): $P_d$ = $P(E_s = short)$, where $E_s$ is an indicator variable (1 if the user prefers a short explanation, 0 otherwise).

\subsection*{Explanation Scope}
Let $P_s$ be a Bernoulli random variable representing the probability that the user prefers explanations of a particular scope (local, global): $P_s$ = $P(E_s = local)$, where $E_s$ is an indicator variable (1 if the user prefers a local explanation, 0 otherwise).

\subsection{Integration of explanation preferences into RDDL}
By modeling human explanation preferences as probabilistic variables in RDDL, the robot can probabilistically reason about these preferences and structure its explanations accordingly. This approach allows the robot to optimize its actions to align with human preferences, enhancing the effectiveness and satisfaction of its explanations.

\subsection*{Transition Function}
\begin{lstlisting}
cstate Fluent
{
  E_r : {textual, visual};
  E_dl : {rich, poor};
  E_d : {long, short};
  E_s : {local, global};
}
\end{lstlisting}

In this example, the RDDL domain file defines the possible states of the human explanation preferences. The action of explaining must incorporate its attributes (representation, detail level, duration, and scope) in their current states. 

\begin{figure}[h]
\centering
\includegraphics[width=0.45\textwidth]{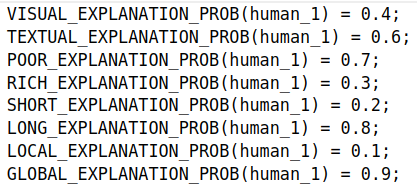}
\caption{Human preferences are represented as probabilities. Probabilities of variables that are values of the random preference variables, e.g., visual and textual for explanation representation, are instantiated such that the sum of their probabilities is 1.0 to have a valid probability distribution.}
\label{fig:pp}
\end{figure}

To approach explaining as task generation, the RDDL problem file specifies the initial state of the system and the probabilities associated with different modalities and their conditional probabilities with representations, detail levels, durations, and scopes (see Figure \ref{fig:pp}). This setup serves as a probabilistic model of human preferences for explanations in human-robot interaction, facilitating the planning and generation of explanations that align with user preferences.

\begin{figure}[h]
\centering
\includegraphics[width=0.45\textwidth]{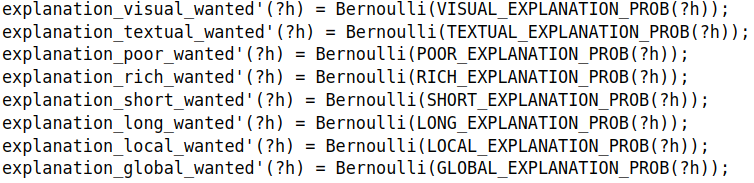}
\caption{Propagation of preference variables through Bernoulli functions. Robots can probabilistically track changing human preferences but also sometimes make mistakes, which more closely mimics real behavior compared to using only deterministic planning.}
\label{fig:dpp}
\end{figure}

In the domain file (see Figure \ref{fig:dpp}), values of preference variables are propagated through states using the Bernoulli function. Such behavior allows the robot to respect human preferences.

\begin{figure}[h]
\centering
\includegraphics[width=0.45\textwidth]{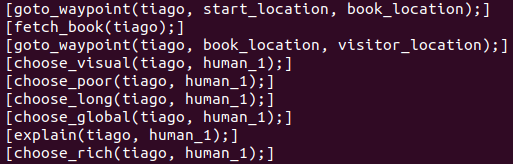}
\caption{Robot librarian starts from start\_location and heads towards book\_location to pick up the book. After picking up the book, it heads towards the visitor's location to hand it to the visitor. Assuming that the robot is too late, it explains its actions to the visitor while choosing visual representation, the poor level of detail, the long duration, and the global scope of its explanation while trying to respect visitor explanation preferences.}
\label{fig:plan}
\end{figure}

Figure \ref{fig:plan} shows the possible plan where human explanation preferences are incorporated.

\section{Discussion}
To customize domain and problem files for different domains, it is enough to do the following:
\begin{enumerate}
\item \textit{Modifying Non-Fluents}: Adjusting the initial and conditional probabilities in the ''non-fluents´´ section of the problem file to reflect different scenarios.

\item \textit{Adjusting Initial State}: Modifying the ''init-state´´ section to set different starting conditions for the human and robot states.
\end{enumerate}

One of the primary contributions of this work is the development of a flexible framework that can adapt to various contexts and user needs. By defining explanation attributes (representation, detail level, duration, and scope) and integrating them into a probabilistic model, we can generate explanations tailored to individual preferences. This level of customization is crucial in diverse application areas ranging from assistive robotics to customer service and educational robots. The RDDL domain and problem files are templates easily modified for different scenarios, showcasing the model’s versatility.

The personalized explanations generated by our model have significant implications for human-robot interaction (HRI). Effective communication is a cornerstone of successful HRI, and explanations that align with user preferences can significantly enhance the user experience. When users receive explanations in a format they find intuitive and timely, their understanding of the robot’s actions improves, leading to increased trust and collaboration.
Moreover, our model’s ability to account for different user states—calm, confused, and stressed—ensures that explanations are personalized and context-aware.

Despite its strengths, our approach does have some limitations. The accuracy of the probabilistic model heavily depends on the quality and representativeness of the initial and conditional probabilities. In real-world applications, these probabilities may need to be fine-tuned based on empirical data, which can be time-consuming and resource-intensive.
Additionally, while RDDL provides a robust framework for modeling, its complexity can be a barrier to those unfamiliar with its syntax and semantics. This might limit the immediate applicability of our approach to practitioners without a background in RDDL or similar formal languages.

\section{Conclusion and Future Work} 
\label{sec:conclusion}
We have presented a novel approach to modeling explanation preferences in human-robot interaction by integrating probability theory and Relational Dynamic Influence Diagram Language (RDDL). Our model encapsulates different attributes that influence user preferences: representation, level of detail, duration, and scope, by employing a probabilistic framework. This framework enables the generation of personalized explanations that cater to the needs and expectations of users interacting with robots.
We believe that personalized explanations of robot navigation are very important factors in social robot navigation.
The better robots can explain their navigation, the better humans can understand them.
Such explanations can be used as input in human-in-the-loop social navigation learning research.
We demonstrated the flexibility and adaptability of our model to various scenarios by providing a generalized RDDL domain and a problem instance. Our model systematically accounts for users' preferences when receiving robot explanations by leveraging conditional probability distributions and utility functions.

Future work will explore the extension of this model to incorporate dynamic user feedback, allowing for real-time adaptation of explanation strategies based on user responses. Additionally, integrating machine learning to refine the probabilistic parameters and utility functions could further enhance the personalization capabilities of the model. For example, reinforcement learning algorithms could be used to adjust the model dynamically as more data about user preferences and states becomes available. This would enable the robot to improve its explanation strategies continuously over time.

Including a wider range of explanation attributes, such as emotional tone and interactivity, could further enrich the explanations and make them more engaging. Exploring the cross-cultural applicability of the model would also be valuable, as preferences for explanations can vary significantly across different cultural contexts. Another promising direction is the integration of real-time user feedback mechanisms. By allowing users to provide immediate input on the explanations they receive, the model can quickly adapt and personalize future explanations even more effectively. This feedback loop would enhance the personalization process and empower users to have greater control over their interactions with robots.

Overall, our research contributes to the growing body of work aimed at making human-robot interaction more intuitive and effective. By focusing on the critical aspect of explanation generation, we have laid the groundwork for developing personalizable intelligent systems that perform tasks efficiently and communicate their actions transparently and understandably. Personalized explanations can lead to improved user satisfaction, a better understanding of robotic actions, and, ultimately, a more harmonious integration of robots into everyday life.



\balance
\bibliographystyle{plainnat}
\bibliography{references}

\end{document}